\documentclass[conference,11pt]{IEEEtran}

\usepackage{amsmath,amssymb,amsfonts}

\usepackage{graphicx}
\usepackage{textcomp}
\usepackage{multirow}
\usepackage{booktabs}
\usepackage{xcolor}
\usepackage{array}
\usepackage{tabularx}
\usepackage{float}
\usepackage{comment}
\usepackage[dvipsnames,svgnames,table, x11names]{xcolor}
\usepackage{caption}
\usepackage[absolute,overlay]{textpos} 

\usepackage{algorithm}
\usepackage{algpseudocode}

\usepackage{hyperref}

\usepackage{tikz}
\usetikzlibrary{shapes.geometric, arrows.meta, positioning, fit, backgrounds, calc}

\usepackage{pifont}

\newcommand{\upstairs}[1]{\textsuperscript{#1}}

\onecolumn

\title{\texttt{ConTrans}: Learning Text-enhanced Local--global Temporal Representations for Zero-shot Temporal Action Localization}

\author{
\IEEEauthorblockN{
Kanchan Keisham\upstairs{1*}, Thenukan Pathmanathan\upstairs{2}, Thangarajah Akilan\upstairs{2*}
}
\IEEEauthorblockA{
\upstairs{1} Vellore Institute of Technology, India\\
\upstairs{2} Lakehead University, Canada\\
kanchan.keisham@vit.ac.in, takilan@lakeheadu.ca
}
}

\usepackage{fancyhdr}
\begin{document}

\maketitle

\begin{textblock*}{21cm}(0cm,0.5cm)
\begin{center}
    {\small\textit{Accepted in The 39th Canadian Conference on Artificial Intelligence (Canadian AI 2026) }}
\end{center}
\end{textblock*}

\begin{abstract}
Zero-shot Temporal Action Localization (ZS-TAL) aims to detect and locate previously unseen actions in untrimmed videos. However, existing approaches primarily focus on modeling long-range contextual information, often neglecting the critical relative-offset-based local correlations between video frames. Furthermore, their performance is hindered by limited feature representation capabilities due to the shallow nature of their network architectures.  In this paper, we address these limitations by introducing a novel local-global multi-scale feature representation module. We propose a novel multi-scale encoder architecture, termed \texttt{ConTrans}, that integrates convolutional (Conv) inductive biases with transformer Self-attention to jointly capture fine-grained local dependencies and long-range global context, leading to more comprehensive feature representations than existing methods.
 Experimental evaluations on the ActivityNet-1.3 and THUMOS14 datasets demonstrate that \texttt{ConTrans} significantly outperforms existing methods, establishing a new benchmark for ZS-TAL.
\end{abstract}



\section{Introduction}
Temporal action localization (TAL) aims to detect and classify actions in long, untrimmed videos. Most existing TAL approaches \cite{shao2023action, zhang2022actionformer} operate under a closed-set assumption, where training and inference share the same action categories. However, real-world applications such as video retrieval and anomaly detection require recognizing unseen actions, motivating zero-shot temporal action localization (ZS-TAL). ZS-TAL enables models to localize actions from novel categories without labeled training examples.
Large-scale vision–language (ViL) models, including CLIP \cite{radford2021learning}, ALIGN \cite{jia2021scaling}, UniCL \cite{yang2022unified}, and ZIM \cite{kim2025zim}, have demonstrated strong zero-shot generalization by aligning visual and textual representations using large-scale web data. Beyond image understanding \cite{luo2025zero, li2025lgd}, these models have significantly improved robustness and generalization in video tasks such as action recognition \cite{bosetti2024text, ye2025zero}, captioning \cite{tewel2022zero, li2025temporal}, and object tracking \cite{tran2024z}. zero-shot inference is typically performed by measuring similarity between visual features and text embeddings representing semantic queries, making ViL representations promising for ZS-TAL.

Recently, zero-shot temporal action detection (ZS-TAD) has gained attention as a related task. STALE \cite{nag2022zero} is an early and influential method that aligns visual and textual embeddings within a one-stage detection framework, preserving foreground information via representation masking and improving classification through text prompt tuning (TPT). Subsequent work \cite{raza2024zero} further explores multi-modal prompt learning to adapt CLIP for temporal detection while reducing computational cost by pretraining prompts on image datasets and freezing them during TAD training.

Despite this progress, existing ZS-TAD methods struggle to capture fine-grained temporal cues and lack effective modeling of contextual interactions between visual and textual modalities, leading to imprecise action boundaries and limited robustness in complex videos. To address these limitations, we propose \texttt{ConTrans}, a novel framework that enhances zero-shot temporal action localization through Text-enhanced temporal representations. \texttt{ConTrans} integrates semantic textual information with visual features via multi-scale fusion and attention mechanisms, capturing both global and local context across modalities. This design improves sensitivity to subtle temporal variations and enables more accurate localization and recognition of unseen actions in open-set video scenarios.

In summary, the main contributions of this work are as follows:
\begin{itemize}
\item \textbf{A novel self-attention model for ZS-TAL:} We propose \texttt{ConTrans}, combining self-attention and Conv to capture long-range temporal dependencies and fine-grained local motion, addressing both contextual reasoning and precise action boundary localization.
\item \textbf{Text-enhanced multi-scale visual–textual fusion:} A hierarchical mechanism aligns visual and textual features across temporal scales, leveraging semantic guidance to improve cross-modal interactions and zero-shot action localization.
\item \textbf {Rich temporal representation for open-set detection:}
Attention-based cross-modal reasoning produces rich temporal features sensitive to subtle boundary variations, enabling accurate detection of unseen actions in complex videos.
\item \textbf{Improved performance in ZS-TAL}:
\texttt{ConTrans} achieves state-of-the-art results on benchmark datasets ActivityNet-1.3~\cite{caba2015activitynet} and THUMOS14~\cite{idrees2017thumos}, demonstrating robust performance in zero-shot temporal action localization.
\end{itemize}

\section{Related Work}
\label{sec:related}

\subsection{Temporal action localization} 

TAL aims to detect and classify action segments in long, untrimmed videos. Existing methods fall into two categories: (i) Two-stage approaches that first generate temporal proposals and then classify them \cite{zhao2021video, su2021bsn++}, and (ii) Single-stage approaches that perform classification and localization in a single forward pass \cite{lin2021learning}. Recent state-of-the-art methods focus on proposal-free models for better efficiency and accuracy. Transformer-based models like Actionformer \cite{zhang2022actionformer} generate multi-scale feature representations while simultaneously classifying actions and detecting boundaries. However, they require full supervision and large datasets, which are often impractical. Inspired by proposal-free methods, we introduce a simple yet effective solution to generalize to unseen actions with minimal training data.

\subsection{Zero-shot temporal action localization} 

In zero-shot settings, the goal is to detect and localize unseen action instances not present in the training data. EffPrompt \cite{ju2022prompting} introduced zero-shot temporal action localization (ZS-TAL) with a two-stage approach, generating action proposals and classifying them using CLIP \cite{radford2021learning}. ZEETAD \cite{phan2024zeetad} employs CLIP to encode RGB frames and action categories, using frame-level similarity to construct semantic representations. However, such a formulation primarily captures single-scale semantics and may struggle with temporally complex or motion-dependent actions. Our method addresses this limitation by integrating visual and textual features at multiple scales, allowing semantic alignment to adapt across varying temporal extents.  However, it is computationally expensive due to redundant proposals. STALE \cite{nag2022zero} offers a single-stage approach with parallel classification and localization, using class-agnostic masking for adaptability but lacking explicit boundary regression. UnLoc \cite{yan2023unloc} builds a feature pyramid to classify actions and detect boundaries at each frame. \cite{raza2024zero} improves cross-modal alignment with multi-modality prompting but relies on a complex two-step training process. GAP \cite{du2024towards} enhances action proposals using static CLIP information but remains inefficient due to its two-stage nature. To address these issues, we propose a one-stage model that efficiently classifies actions and detects temporal boundaries.

\subsection{Vision Language }

Vision-language models (VLM) combine computer vision and natural language processing to address tasks like image-text retrieval \cite{wang2019camp} and visual question-answering \cite{antol2015vqa}. Pre-trained on large-scale image-text pairs, VLMs can be applied to visual recognition tasks without additional fine-tuning. CLIP \cite{radford2021learning}, a large-scale VLM trained with an image-text contrastive objective, has shown outstanding zero-shot performance \cite{yang2023multicapclip}. Recently, CLIP has been adapted for video tasks such as text-based action localization \cite{miech2020end}, typically using a two-stage approach involving foreground cropping and alignment. However, this process suffers from error propagation, where mistakes in the first stage affect the second. To overcome this, we propose a single-stage model for simultaneous classification and localization without additional CLIP fine-tuning.
\section{Proposed method}
\label{sec:method}
\begin{figure*}[ht!]
\begin{center}
   \includegraphics[trim={2.5cm, 2cm, 2.2cm, 2.0cm}, clip, width=1.0\linewidth]{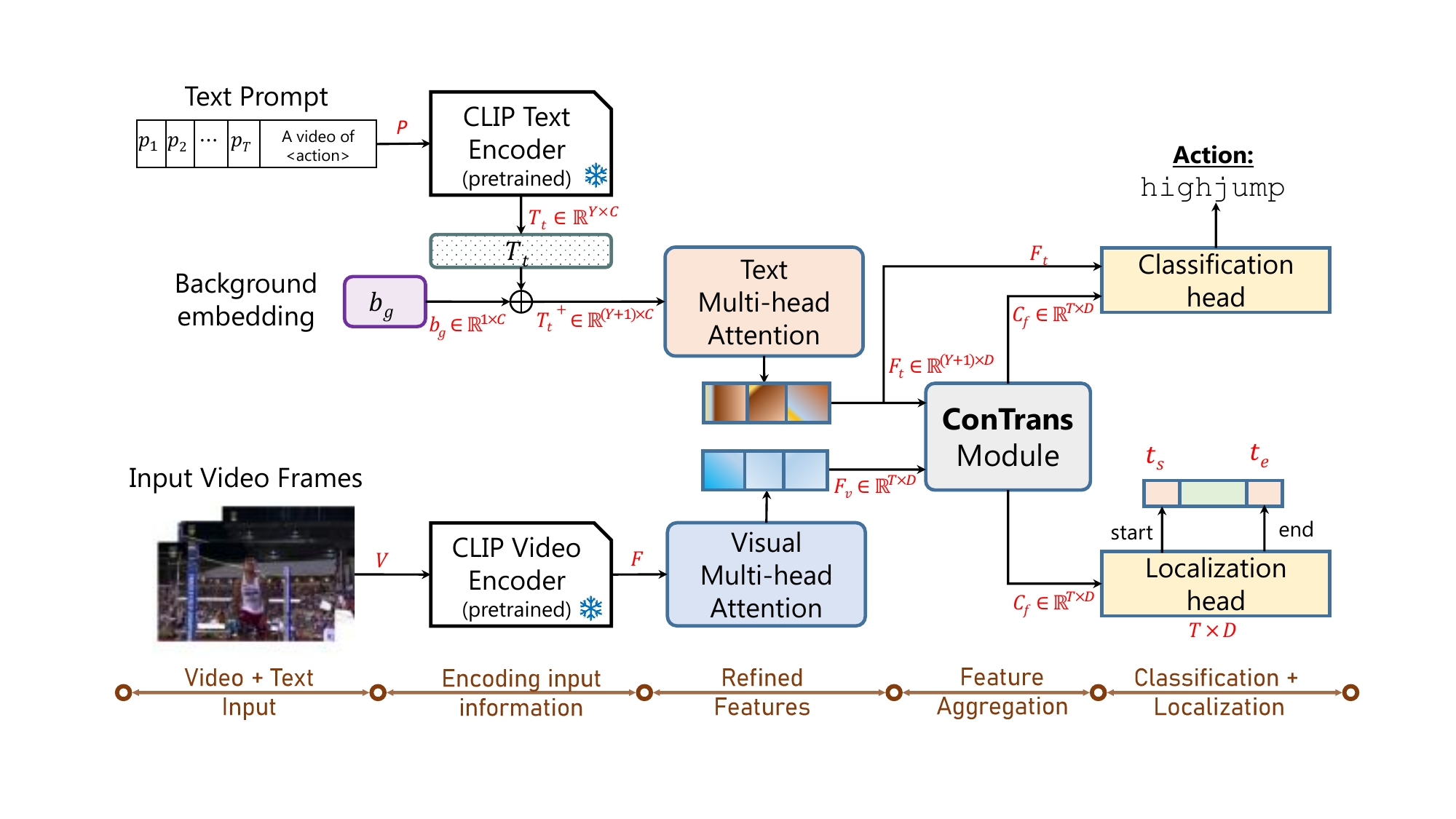}
\end{center}
\vspace{-0.2cm}
   \caption{Overview of the proposed \texttt{ConTrans} architecture.}
\label{main}
\vspace{-0.2cm}
\end{figure*}

This section provides a comprehensive overview of our proposed \texttt{ConTrans} model, with the overall framework illustrated in Fig. \ref{main}. We begin by defining the problem in Subsection 3.1. Next, Subsection 3.2 discusses how the pre-trained CLIP model is utilized as an embedding module for ZS-TAL. Finally, Subsection 3.3 delves into the proposed local-global multi-scale feature representation module, detailing its role in action classification and localization.

\subsection{Problem definition}

The main goal of the proposed method is to identify temporal action instances in unseen videos, defined by their start time, end time, and action label.
For instance, in the dataset used in ZS-TAD, each untrimmed video $V$, consisting of $S$ snippets, is associated with a set of action annotations $L = \{(t_s^i, t_e^i, y^i)\}_{i=1}^{S}$, where $t_s^i$ and $t_e^i$ denote the start and end times of the $i$-th action instance, and $y^i \in Y$ is its action label.
Both closed-set and open-set scenarios are examined. In the closed-set scenario, the training and evaluation action labels are identical (\(Y_{train} = Y_{val}\)), whereas in the open-set scenario, the action labels for training and evaluation are mutually exclusive (\(Y_{train} \cap Y_{val} = \emptyset\)).

\subsection{Pre-trained CLIP as encoding module}

Recent studies \cite{nag2022zero} leverage CLIP as the backbone for ZS-TAD due to its strong zero-shot transfer capability. CLIP is pre-trained on 400 million image–text pairs and maps visual and textual inputs into a shared latent space using image and text encoders. For ZS-TAD, we use the pre-trained CLIP encoders to extract visual and textual features, keeping the model frozen during training.
For visual encoding, we sample $T$ consecutive frames from a video $V = \{{l_{1}, l_{2}, \cdots, l_{T}}\}$. 
The CLIP image encoder extracts visual features $F \in \mathbb{R}^{T \times D}$, where $D$ is the feature dimension and $T$ is the number of sampled frames. To capture global temporal context, we apply a multi-head attention (MHA) mechanism \cite{vaswani2017attention} with $(query, key, value) = (F, F, F)$, producing refined visual features $F_v$ as shown in Eq.~\eqref{eq:mha}.
\begin{equation}
\begin{gathered}
\label{eq:mha}
F_{v} = \mathrm{MHA}(F,F,F) \in R^{T \times D}.
\end{gathered}
\end{equation}
\noindent\textbf{Textual Encoding--}
Each action category $y_i \in Y$ is represented using a template, ``\texttt{a video of <action>}'', where \texttt{<action>} is replaced by the class label $y_i$. This text prompt, $P_T$, is fed into the pre-trained CLIP text encoder to obtain textual embeddings. Let $T_t \in \mathbb{R}^{Y \times C}$ denote the resulting embeddings for all $Y$ action classes, where $C$ is the embedding dimension. Since temporal action localization also requires background classification, we introduce a learnable background embedding $b_g \in \mathbb{R}^{1\times C}$ and append it to $T_t$, as in Eq.~\eqref{eq:text}:
\begin{equation}
\label{eq:text}
T_t^{+} = [T_t ; b_g] \in \mathbb{R}^{(Y+1) \times C}.
\end{equation}

The refined textual features $F_t$ are then obtained using a multi-head attention mechanism, as defined in Eq.~\eqref{eq:prompt}. Positional encoding (PE) is excluded from both visual and textual branches, as our ablation study indicates that it degrades performance.
\begin{equation}
\label{eq:prompt}
F_t = \text{MHA}(T_t^{+}, T_t^{+}, T_t^{+}) \in \mathbb{R}^{(Y+1) \times D}.
\end{equation}
\subsection{{ConTrans} module}

\begin{figure*}[bp!]
\centering
   \includegraphics[trim={5.0cm, 7.6cm, 16cm, 2cm}, clip, width=0.65\linewidth]{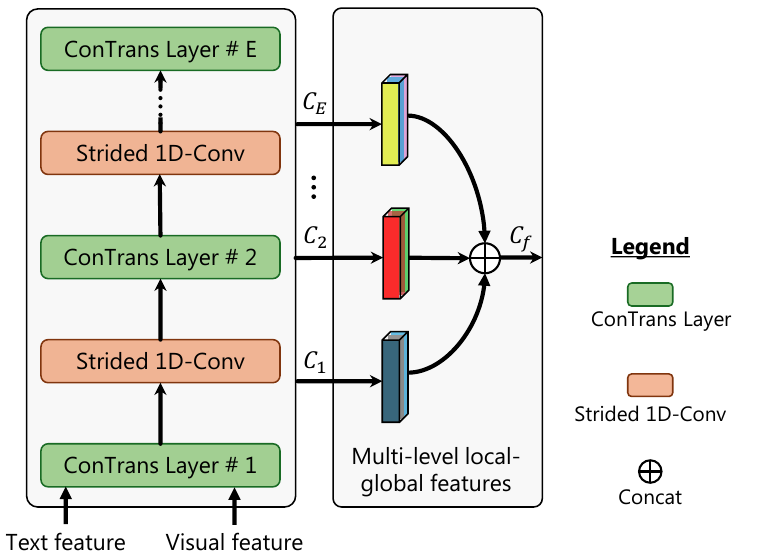}
   \vspace{-0.2cm}
   \caption{The proposed \texttt{ConTrans} module, which comprises ``E'' number of \texttt{ConTrans} layers, with each layer followed by a strided 1D-Conv for downsampling. }
\label{cons}
\vspace{-0.2cm}
\end{figure*}

\begin{figure*}[tp!]
\centering
   \includegraphics[trim={15cm, 7.35cm, 7cm, 1.0cm}, clip, width=0.62\linewidth]{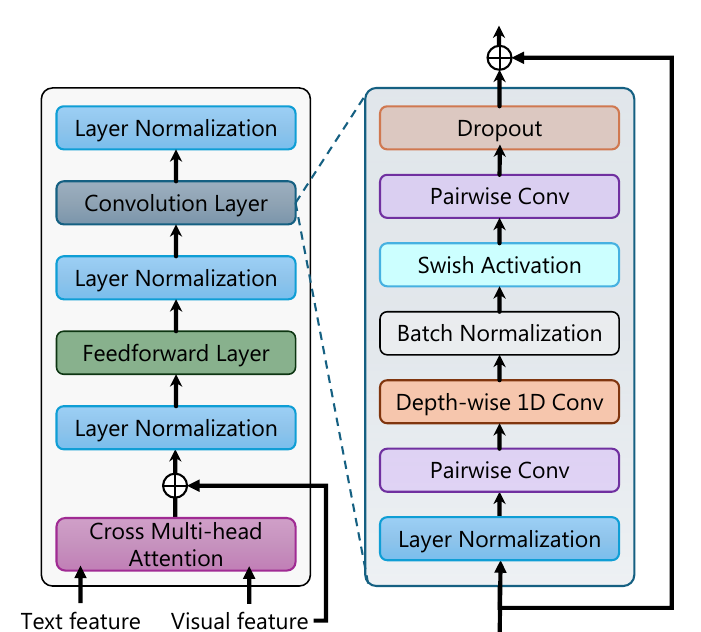}
   \caption{A detailed view of the \texttt{ConTrans} layer. It captures both local and global context via a combination of cross-attention and convolutional layers. }
\label{conlayer} \vspace{-0.4cm}
\end{figure*}

Fig.~\ref{cons} illustrates the high-level structure of the proposed \texttt{ConTrans} module, while Fig.~\ref{conlayer} provides a detailed connectivity diagram of a \texttt{ConTrans} layer. 
The \texttt{ConTrans} module consists of multiple stacked \texttt{ConTrans} layers followed by strided depthwise 1D Conv layers. Each \texttt{ConTrans} layer is designed to capture both local and global cross-modal interactions between textual and visual features at multiple hierarchical levels. This hierarchical design progressively refines cross-modal representations, enabling accurate detection of actions across varying temporal durations.
As shown in Fig.~\ref{conlayer}, each \texttt{ConTrans} layer comprises a cross multi-head attention module, followed by a Conv layer and a feed-forward network, with layer normalization applied throughout to ensure training stability. Given visual features $F_v$ and textual features $F_t$, the model uses contextual visual information to guide textual representations in identifying informative temporal regions. Specifically, text features act as queries, allowing the model to focus on video regions most relevant to each action category. This design is especially effective in zero-shot settings, where textual semantics provide critical guidance for localizing unseen actions. To this end, we employ cross-attention with $(query, key, value) = (F_t, F_v, F_v)$, as in Eq.~\eqref{eq:weights}.

\begin{equation}
\begin{aligned}
\label{eq:weights}
q &= W_q \,\mathcal{LN}(F_t) \in \mathbb{R}^{(Y+1) \times d}, \\
k &= W_k \,\mathcal{LN}(F_v) \in \mathbb{R}^{T \times d}, \\
v &= W_v \,\mathcal{LN}(F_v) \in \mathbb{R}^{T \times d},
\end{aligned}
\end{equation}
where $W_q, W_k, W_v \in \mathbb{R}^{D \times d}$ are learnable projection matrices. The output of the multi-head cross-attention mechanism, followed by a feed-forward layer, $FF(\cdot)$, and layer normalization, $\mathcal{LN}(\cdot)$, is defined in Eq.~\eqref{eq:scalar}.
\begin{equation}
\begin{gathered}
\label{eq:scalar}
\mathrm{MHA}(q,k,v) = \text{Concat} (head_1, head_2, \cdots, head_h); \quad
head_i = \text{Attention}(q,k,v) \\
C_a = \mathcal{LN}(FF(\mathrm{MHA})),
\end{gathered}
\end{equation}
where $C_a\in \mathbb{R}^{(Y+1) \times d}$ is the final cross-attention output. 
To model local cross-modal correlations, $C_a$ is first projected into temporal dimension $T\times d$ and processed by a Conv block. This block complements the global context captured by attention by learning fine-grained local dependencies. The Conv output is combined with a residual connection and normalized to generate $C_{o}$, as expressed in Eq.~\eqref{eq:conv}.
\begin{equation}
\begin{gathered}
\label{eq:conv}
C_{o} = \mathcal{LN}(C_a + \text{Conv}(C_a)).
\end{gathered}
\end{equation}

Inspired by \cite{gulati2020conformer}, the Conv block consists of a pointwise Conv with a gated linear unit (GLU), followed by a 1D depthwise Conv and batch normalization (see Fig.~\ref{conlayer}). The resulting $C_{o}$ captures both local and global cross-modal context at each \texttt{ConTrans} layer. To obtain rich multi-scale temporal representations, the output of each \texttt{ConTrans} layer is downsampled using 2× strided depthwise 1D Conv \cite{zhang2022actionformer} before being passed to subsequent layers as: 
\begin{equation}
\begin{gathered}
\label{eq:down}
C\hat{_o} = C_o(\downarrow); \quad
C\hat{_e} = \mathcal{LN}(FF(\mathrm{MHA}(C\hat{_o}))), \text{~and}\\
C\hat{_{e_{i}}} = \mathcal{LN}((C\hat{_e} + \text{Conv}(C\hat{_e})),    i = \{1,\cdot \cdot \cdot, E\}.
\end{gathered}
\end{equation}
Thus, $\{C_{e}, C_{e_{1}},C_{e_{2}},\cdot \cdot \cdot C_{E}\}$ represents the output from $E$ \texttt{ConTrans} layers. To enable the model to leverage both low-level and high-level representations, we then concatenate the multi-scale outputs as shown in Eq.~\eqref{eq:concat}.
\begin{equation}
\begin{gathered}\label{eq:concat}
C_{f} = \text{Concat}(C_{e}, C_{e_{1}},C_{e_{2}},\cdots, C_{E}).
\end{gathered}
\end{equation}
For action classification, the class probability  $Prob$ is computed by the dot product between $C_f$   and the refined text features $F_t$ as defined in Eq \eqref{eq:prob}.  
\begin{equation}
\begin{gathered}\label{eq:prob}
Prob = \text{Softmax}(F_t C_f^T)
\end{gathered}
\end{equation}
where $Prob_{i,j}$ denotes the probability of the $i$-th class at the the $j$-th temporal snippet.
For temporal localization, we follow prior work \cite{nag2022zero, raza2024zero} and predict 1D action masks over the full video duration. A stack of three 1D-Conv layers operates on $C_f$ to produce foreground probabilities $A_L \in \mathbb{R}^{T \times 1}$, where each element represents the likelihood of an action at the corresponding temporal snippet as shown in Eq \eqref{eq:head}.
\begin{equation}
\begin{gathered}\label{eq:head}
 A_L= \text{Sigmoid}(\text{1D-CNN}(C_{f})).
\end{gathered}
\end{equation}
In addition, the model indicates the presence and strength of action at each snippet by generating a confidence score $Conf$ using a series of Conv layers, as shown in Eq \eqref{eq:local}. Higher confidence values facilitate more accurate boundary estimation,
\begin{equation}
\begin{gathered}\label{eq:local}
 Conf= \text{ReLU}(\text{1D-CNN}(C_{f})).
\end{gathered}
\end{equation}
\section{Training objective and Inference}

\label{sec:training}
For label assignment, the ground-truth annotations are structured as follows. Given a training video with annotated temporal intervals and class labels, all snippets within the duration of an action instance are assigned the same action class, while snippets outside any action interval are labeled as background. For each action snippet in the class stream, we assign a binary instance mask in the action mask stream that spans the entire video length at the corresponding snippet index. All snippets belonging to the same action instance share the same instance-specific mask.
For action classification, we employ the cross-entropy loss $L_c$ between the predicted class probabilities $p \in Prob$ and the ground-truth labels $y \in \mathbb{R}^{(Y+1)\times C}$ for each action snippet, as defined in \eqref{eq:entropy}.
\begin{equation}
\begin{gathered}\label{eq:entropy}
 L_c = \text{CrossEntropy}(p,y).
\end{gathered}
\end{equation}
For the action localization loss, binary dice loss \cite{nag2022zero} along with weighted cross-entropy loss is computed between the predicted action mask $A \in R ^{(TX1)}$ of the $i-th$ action snippet and the ground-truth action mask $g\in R ^{(TX1)} $ as defined in \eqref{eq:mask}. 
\begin{equation}
\begin{gathered}\label{eq:mask}
M= \beta_f \sum_{i=1}^T g(t)log(A(t)) + \beta_b \sum_{i=1}^T (1-g(t))(1-log(A(t))) \\ 
+ \lambda\left(1-\frac{A^Tg}{\sum_{i=1}^T (A(t))^2 + (g(t))^2}\right),
\end{gathered}
\end{equation}
where $\beta_f$, $\beta_b$ are the inverse of the foreground/background snippet's proportion and $\lambda\!\!=\!\!0.4$ is the loss trade-off coefficient. For the generated confidence score, we adopt $L2$ loss to calculate regression {error} and set the weight term $\lambda\!\!=\!\!10$. 
Thus, the overall loss is as follows: 
\begin{equation}
\begin{gathered}\label{eq:total}
 Loss= L_c + M + \lambda \times L2.
\end{gathered}
\end{equation}

During testing, action instance predictions for each test video are generated using the classification predictions \(Prob\) and the mask predictions \(A\). For \(Prob\), we only consider snippets where the class probabilities surpass a threshold \(\theta_c\) and select the highest-scoring snippets. For each of these high-scoring action snippets, the corresponding temporal mask is obtained by applying a threshold to the \(t_i\)-th column of \(A\) using the localization threshold \(\Theta\). To ensure a sufficient number of candidates, we use a set of thresholds \(\Theta = \{\theta_i\}\). For these candidate snippets, the confidence score \(s\) is calculated by multiplying the classification score by the maximum mask score. Finally, SoftNMS \cite{bodla2017soft} is applied to obtain the top-ranked results.
The proposed method is evaluated using two standard ZS-TAL datasets: ActivityNet-1.3 \cite{caba2015activitynet}, and THUMOS14 \cite{idrees2017thumos}. ActivityNet-1.3 consists of 200 action categories and a total of 19,994 videos. In accordance with the standard protocol, the dataset is split into training, validation, and test sets in a 2:1:1 ratio.  THUMOS14 contains 20 action categories with 200 validation videos and 213 test videos. Both datasets include annotations for temporal boundaries and corresponding action labels.

\subsection{Implementation details} 

For fair comparison with prior works \cite{nag2022zero, raza2024zero, yan2023unloc}, we adopt a pre-trained CLIP (ViT-B/16) visual and text encoders with feature dimension $D\!=\!512$, keeping them frozen during training. The encoder, i.e., ViT-B/16 + Transformer, is used to generate textual embeddings. Video frames are resized to $224\!\!\times\!\!224$, and 77 textual tokens are used for both datasets \cite{nag2022zero}. Extracted features are temporally rescaled to 100 and 256 for ActivityNet-1.3 and THUMOS14, respectively.
The model employs 4 \texttt{ConTrans} blocks with 8 attention heads and is trained for 9 epochs using the Adam optimizer, with learning rates of $10^{-4}$ for ActivityNet-1.3 and $10^{-5}$ for THUMOS14. Performance is evaluated using the standard metric mean average precision (mAP) at multiple IoU thresholds. 

\section{Comparison results with the state-of-the-art models}

\begin{table*}[tp!]
\centering

\begin{tabular}{l|l|llll|llllll}
\hline
Data split & \multirow{2}{*}{Model} & \multicolumn{4}{c|}{ActivityNet-1.3}  
 & \multicolumn{6}{c}{THUMOS14}    \\ \cline{3-12}
Train:Test &  & 0.5 & 0.75 & 0.95 & mean  & 0.3 & 0.4 &0.5 & 0.6 & 0.7 &mean \\
\hline
   \multirow{8}{*}{75:25}  &B-II  &32.6 &18.5 &5.8 &19.6  &28.5 &20.3 &17.1 &10.5 &6.9 &16.6\\
    
                         &B-I &35.6 &20.4 &2.1 &20.2 &33.0 &25.5 &18.3 &11.6 &5.7 &18.8\\
   
                         &EffPrompt &37.6 &22.9 &3.8 &23.1  &39.7 &31.6 &23.0 &14.9 &7.5 &23.3\\
                    
                         &mProTEA &44.5 &27.4 &7.9 &27.6 &43.1 &38.2 &28.2 &18.1 &8.7 &27.9\\
                         &STALE &38.2 &25.2 &6.0 &24.9 &40.5 &32.3 &23.5 &15.3 &7.6 &23.8 \\
                         &GAP &47.6 &32.5 &8.6 &31.8 &52.3 &44.2 &32.8 &22.4 &12.6 &32.9\\
                        
                         &GRIZAL &46.4 &32.5 &6.8 &30.1 &43.2 &- &25.7 &- &9.8 &27.0 \\
                        \rowcolor{green!15} 
                        \cellcolor{white}  & Ours &\textbf{51.9} &\textbf{33.1} &\textbf{12.2} &\textbf{35.2}  &\textbf{51.3} &\textbf{45.6} &\textbf{33.1} &\textbf{22.7} &\textbf{12.9} &\textbf{33.3}\\
\hline
 \multirow{8}{*}{50:50}  &B-II  &32.1 &20.7 &3.7 &12.9 &21.0 &16.4 &11.2 &6.3 &3.2 &11.6\\
    
                         &B-I &28.0 &16.4 &1.2 &16.0 &27.2 &21.3 &15.3 &9.7 &4.8 &15.7 \\
   
                         &EffPrompt &32.0 &19.3 &2.9 &19.6 &37.2 &29.6 &21.6 &14.0 &7.2 &21.9\\
                        
                         &mProTEA &41.8 &24.6 &6.1 &25.6 &41.2 &36.3 &26.3 &16.8 &8.4 &26.1\\
                         &STALE   &32.1 &20.7 &5.9 &20.5 &38.3 &30.7 &21.2 &13.8 &7.0 &22.2 \\
                         &GAP  &41.6 &26.2 &6.1 &26.4 &44.2 &36.0 &27.1 &15.1 &8.0 &26.1\\
                         &GRIZAL &39.9 &25.7 &6.6 &25.7 &40.0 &- &25.0 &- &9.1 &25.2 \\
                        &UnLoc &43.7 &- &- &-&-&-&-& -  &-&- \\
                        \rowcolor{green!15} 
                        \cellcolor{white}  
                         &Ours &\textbf{45.3} &\textbf{30.3} &\textbf{10.5} &\textbf{32.5}   &\textbf{44.5} &\textbf{36.3} &\textbf{26.8} &\textbf{15.7} &\textbf{8.8} &\textbf{27.2}\\ \hline
\end{tabular}
\caption{Comparison of the proposed model with other state-of-the-art models on the ActivityNet-v1.3~\cite{caba2015activitynet}, and THUMOS14~\cite{idrees2017thumos} datasets for zero-shot Temporal Action Localization (ZS-TAL), evaluating the mean average precision (mAP) at different temporal Intersection over Union (tIoU) thresholds for open-set settings.}
\label{main_table} \vspace{-0.2cm}
\end{table*} 

\begin{table*}[tp!]
\centering

\setlength{\tabcolsep}{4.7pt}
\begin{tabular}{l|l|llll|llllll}
\hline
\multirow{2}{*}{Model} & \multirow{2}{*}{Mode} & \multicolumn{4}{c}{ActivityNet-1.3} & \multicolumn{6}{c}{THUMOS14}    \\ \cline{3-12}
 &  & 0.5 & 0.75 & 0.95 & mean &0.3 &0.4 &0.5 &0.6 &0.7 & mean \\
\hline
 A2Net+I3D  & RGB  &39.6 &25.7 &2.8 &24.8 &45.0 &40.5 &31.3 &19.9 &10.0 &29.3 \\
 B-I+CLIP  & RGB & 28.2 &18.3 &3.7 &18.2 &36.3 &31.9 &25.4 &17.8 &10.4 & 24.3\\
 B-II+CLIP & RGB   &51.5 &33.3 &6.6 &32.7 &57.1 &49.1 &40.4 &31.2 &23.1  &40.2\\
 EffPrompt+CLIP  &RGB  &44.0 &27.0 &5.1 &27.3 &50.8 &44.1 &35.8 &25.7 &15.7 &34.5 \\
 STALE+CLIP & RGB  &54.3 &34.0 &7.7 &34.3 &60.6 &53.2 &44.6 &36.8 &26.7 &44.4\\
 \rowcolor{green!15} 
 \texttt{ConTrans}+CLIP  &RGB    &\textbf{54.5} &\textbf{41.8} &\textbf{15.5} &\textbf{36.6} &\textbf{62.1} &\textbf{53.5} &\textbf{45.1} &\textbf{36.9} &\textbf{27.2} &\textbf{45.1}\\
 \hline
TALNet+I3D  &RGB+Flow &38.2 &18.3 &1.3 &20.2 &53.3 &48.5 &42.8 &33.8 &20.8 &39.8 \\
MUSES+I3D &RGB+Flow  &50.0 &34.9 &6.5 &34.0 &68.9 &64.0 &57.1 &46.7 &31.2 &52.9 \\
B-III+I3D  &RGB+Flow  &47.2 &30.7 &8.6 &30.8 &68.3 &62.3 &51.9 &38.8 &23.7 &- \\
STALE+I3D  &RGB+Flow  &56.5 &36.7 &9.5 &36.4 &68.9 &64.1 &57.1 &46.7 &31.2 &52.9\\
\rowcolor{green!15} 
\texttt{ConTrans}+I3D &RGB+Flow  &\textbf{60.2} &\textbf{42.4} &\textbf{22.3} &\textbf{38.5} &\textbf{69.1} &\textbf{64.6} &\textbf{57.4} &\textbf{47.1} &\textbf{31.5} &\textbf{53.2}\\
\hline
\end{tabular}
\caption{Comparison of the proposed model with other state-of-the-art models using I3D or CLIP encoder backbones on the ActivityNet1.3 and THUMOS14 for zero-shot Temporal Action Localization (ZS-TAL), evaluating the mean average precision (mAP) at different temporal Intersection over Union (tIoU) thresholds on closed-set settings.}
\label{anet_table}\vspace{-0.2cm}
\end{table*}

\subsection{Experimental setup and analysis}
\label{sec:experiments}

\textbf{a. Zero-shot settings: } The performance of the proposed model is assessed in open-set scenarios where $Y_{train} \cap Y_{val} = \emptyset$, meaning the training and evaluation labels do not overlap. We adhere to the evaluation settings and dataset splits outlined in \cite{nag2022zero}. Specifically, the two datasets, ActivityNet-1.3 and THUMOS14, are evaluated under two conditions: 1) training on 75\% of the action categories and testing on the remaining 25\%, and 2) training on 50\% of the action categories and testing on the other 50\%. For consistent and reliable evaluation, we average results across 10 random splits and compare only with methods that adhere to the same evaluation procedure.\\
\textbf{\textit{Competitors--}}
As ZS-TAD is a relatively new problem, only a few competitive methods \cite{nag2022zero,yan2023unloc, raza2024zero}, and GAP \cite{du2024towards} are available for fair comparison. We additionally introduce two CLIP-based baselines: B-I, a two-stage approach combining BMN \cite{lin2019bmn} with CLIP, and B-II, a one-stage TAD model integrated with CLIP. Both baselines use the same CLIP-initialized text encoder. Direct comparison with ZS-TAD \cite{zhang2020zstad} is not feasible due to unavailable code and inconsistent data splits with \cite{ju2022prompting}. To ensure statistical reliability, we perform 10 random category samplings for each configuration, following \cite{nag2022zero}.\\
\textbf{\textit{Performance--}} 
Referring to Table~\ref{main_table},  
it is found that our approach achieves the highest mAP, reaching 35.2 and 33.3 under the 75\%–25\% split, and continues to outperform state-of-the-art methods under the more challenging 50\%–50\% split with 32.5 and 27.2 mAP, respectively. The consistent gains across splits demonstrate strong robustness to limited training data and effective generalization to unseen classes. Performance improvements on ActivityNet-1.3 stem from modeling long temporal segments, while gains on THUMOS14 arise from capturing local–global context for densely occurring short actions.\\

\begin{figure*}[tp!]
\centering
   \includegraphics[trim={0.5cm, 2.5cm, 0.5cm, 1cm}, clip, width=1.0\linewidth]{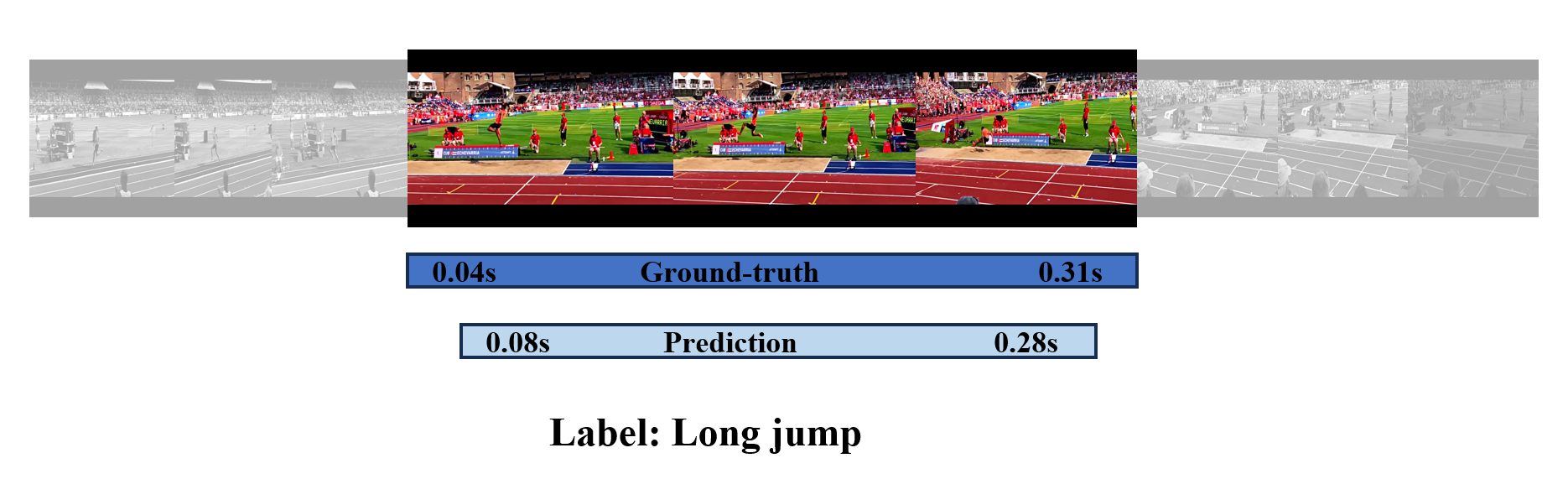}
   Label: \texttt{Long Jump} \vspace{-0.2cm}
   \caption{Qualitative analysis for action class ``Long jump''.}
\label{anet} \vspace{-0.2cm}
\end{figure*}

 \textbf{b. Closed-set settings: } In closed-set scenarios, the action labels used for training and evaluation are identical (i.e., \( Y_{train} = Y_{val} \)). To ensure a fair comparison, we adopt the same dataset splits as reported in previous studies. \\
\textbf{\textit{Competitors--}} The proposed model's performance is evaluated against the latest state-of-the-art approaches, such as STALE \cite{nag2022zero}. Furthermore, we consider several temporal action localization(TAL) methods that leverage the CLIP backbone, as well as three baseline models for comparison: the two-stage CLIP-based baseline B-I, and two one-stage baselines—CLIP-based B-II and B-III, which utilize Kinetics pre-trained I3D. \\
\textbf{\textit{Performance--}} 
As shown in Table \ref{anet_table}, the proposed model improves consistently with more training samples on both ActivityNet-1.3 and THUMOS14. \texttt{ConTrans} outperforms existing methods across different input modalities (RGB and RGB+Flow) and encoder backbones (I3D and CLIP), highlighting its effectiveness across diverse feature representations. Its multi-level feature design captures both fine-grained and high-level temporal context, enabling robust detection of actions with varying durations and strong generalization across datasets and action categories. \\
\textbf{\textit{Computational complexity--}} As demonstrated in Table \ref{complexity}, our model exhibits faster performance in both training and inference times compared to prior SOTA methods. Its efficiency is further highlighted by its early convergence, achieving optimal results in just 9 epochs.\\
\textbf{\textit{Qualitative analysis--}}
The qualitative analysis of the proposed model is presented in Fig.~\ref{anet}. For the action class \emph{Long jump}, the model accurately predicts the temporal boundaries in close alignment with the ground truth. As shown in the figure, the video spans a total duration of 31 seconds [0.04-0.31]s, and the proposed model predicts comparable boundaries at [0.08-0.28]s.

\begin{table}[tp!]
\centering
\begin{minipage}{0.5\linewidth}
    \centering
    
    \begin{tabular}{l|ll}
        \hline
        Model & Train time (hrs) & Infer time (s) \\
        \hline
        EffPrompt~\cite{ju2022prompting} & 3.24 & 135 \\
        STALE~\cite{nag2022zero} & 3.96 & 151.8 \\
        mProTEA~\cite{raza2024zero} & 1.42 & 162.5 \\
        \rowcolor{green!15}
        Ours & \textbf{1.3} & \textbf{31} \\
        \hline
    \end{tabular}
    
    \captionsetup{width=\linewidth}
    \caption{Complexity analysis of \texttt{ConTrans}.}
    \label{complexity}
    
\end{minipage}
\end{table}

\section{Ablation study}\label{sec:ablation}

To assess the efficacy of the proposed approach, we perform extensive ablation studies on the ActivityNet-1.3 dataset using the 75\%-25\% split configuration.    

\textbf{a. Impact of local-global context dependencies:} It can be observed from Table~\ref{local} that removing either the local or global context information results in a performance drop of 2.3\%, highlighting the importance of incorporating both local and global temporal dependencies. Interestingly, the model's performance with either local attention or global attention alone is similar, with only a slight increase of 0.2\% when local attention is used, demonstrating that both forms of attention contribute equally in capturing essential temporal relationships for ZS-TAL.

    \begin{table}[!htb]
       
          \centering
           
              \begin{tabular}{l|llll}
                 \hline
            Model & 0.5 & 0.75 &0.95 &mean   \\
    \hline
       w/o local &49.8 &29.3 &9.8 &32.7\\
       w/o global &50.1 &30.3 &9.8 &32.9\\
       \rowcolor{green!15} 
        full &\textbf{51.9} &\textbf{33.1} &\textbf{12.2} &\textbf{35.2} \\
        \hline
            \end{tabular}
            \caption{Impact of local-global dependencies, wrt. $\text{mAP}_{0.5 - 0.95}$ (\%).}
    \label{local}
             
    \end{table}

\textbf{b. Impact of \texttt{ConTrans} layers: }
We also evaluate the model's performance based on the number of \texttt{ConTrans} layers used to generate multi-scale features. It can be seen from Table \ref{layer} that the model performs optimally with four \texttt{ConTrans} blocks. However, increasing the number of \texttt{ConTrans} blocks results in a rise in model parameters, leading to higher computational costs and the potential for overfitting.

\begin{table}[!tb]
\centering
\begin{minipage}{0.45\linewidth}
    \centering
    
    \begin{tabular}{c|cccc}
        \hline
        Number of layers & ${0.5}$ & ${0.75}$ & ${0.95}$ & mean \\
        \hline
        1 & 50.2 & 30.4 & 10.6 & 33.4 \\
        3 & 51.3 & 32.1 & 10.0 & 34.1 \\
          \rowcolor{green!15} 
        4 & \textbf{51.9} & \textbf{33.1} & \textbf{12.2} & \textbf{35.2} \\
        \hline
    \end{tabular}
    \captionsetup{width=\linewidth}  
    \caption{The impact of the number of \texttt{ConTrans} layers, wrt. $\text{mAP}_{0.5 - 0.95}$ (\%).}
    \label{layer}
    
\end{minipage}
\hfill
\begin{minipage}{0.45\linewidth}
    \centering
    
    \begin{tabular}{l|llll}
        \hline
        Model & 0.5 & 0.75& 0.95 & mean \\
        \hline
        with PE & 50.9 & 31.0 & 9.5 & 33.3 \\
          \rowcolor{green!15} 
        w/o PE & \textbf{51.9} & \textbf{33.1} & \textbf{12.2} & \textbf{35.2} \\
        \hline
    \end{tabular}
    \captionsetup{width=\linewidth}  
    \caption{The impact of PEs on both the visual and text embeddings, wrt. $\text{mAP}_{0.5 - 0.95}$ (\%).}
    \label{pos}
    
\end{minipage}
\end{table}

\textbf{c. Effect of positional encodings on visual and text encoders: } 
We examine the effect of positional encodings on visual and textual encoders. As shown in Table \ref{pos}, adding them reduces performance, likely because dense frame sampling and cross-modal alignment in zero-shot TAD make explicit positional biases detrimental. Consequently, positional encodings are excluded in our model.

\begin{table}[!tb]
\centering
\begin{minipage}{0.5\linewidth}
    \centering
    
    \begin{tabular}{>{\centering\arraybackslash}p{2cm}|
>{\centering\arraybackslash}p{2cm}|
>{\centering\arraybackslash}m{0.5cm}
>{\centering\arraybackslash}m{0.5cm}}
    \hline
     Text  & Video  & \multicolumn{2}{c}{mAP (\%)}  \\
    
    encoder & encoder & 0.5 &mean \\
    \hline
        \ding{55} &\ding{51} &47.1 &32.6 \\
        \ding{51} &\ding{55}  &49.8 &33.1 \\
          \rowcolor{green!15} 
        \ding{51} &\ding{51}   &\textbf{51.9}  &\textbf{35.2} \\
        \hline
        
    \end{tabular}  
    \captionsetup{width=1.0\linewidth}  
    \caption{Effect of encoders in the proposed model.}
    \label{encoder}
    
\end{minipage}
\hfill
\begin{minipage}{0.45\linewidth}
    \centering
    
    \begin{tabular}{>{\centering\arraybackslash}m{1cm}|
>{\centering\arraybackslash}m{1cm}|
>{\centering\arraybackslash}m{1cm}|
>{\centering\arraybackslash}m{1.5cm}}
    \hline
     mAP(\%) & Max pooling &Avg pooling & \cellcolor{green!15} Strided 1D-Conv  \\
    \hline
      0.5 &44.4 &45.4 &   \cellcolor{green!15}\textbf{51.9} \\
    mean   &32.6 &33.1 & \cellcolor{green!15} \textbf{35.2} \\
    \hline
        
    \end{tabular}
    \captionsetup{width=\linewidth}  
    \caption{Effect of different down-samplers.}
    \label{pool}

\end{minipage}
\end{table}

\textbf{d. Effect of frame and text encoders: } 
In this work, text is encoded using the pre-trained CLIP text encoder, and video frames are processed with Transformer multi-head attention. Table \ref{encoder} shows performance drops when either encoder is removed, highlighting their crucial role in the model’s effectiveness.

\textbf{e. Effects of different downsampling methods: } The ablation study at Table \ref{pool} on different downsampling methods for the feature pyramid shows that the strided 1D-CNN achieves the best performance, while max pooling performs the worst, due to the loss of contextual information from aggressively selecting only maximum values.

\section{Conclusion} \label{sec:conclusion}

We propose \texttt{ConTrans}, a framework combining self-attention and convolution to capture both fine-grained local patterns and long-range temporal dependencies. Its multi-scale visual–textual fusion aligns semantic and visual features across temporal resolutions, enabling effective zero-shot action localization. Experiments on ActivityNet-1.3 and THUMOS14 show \texttt{ConTrans} consistently outperforms state-of-the-art methods, demonstrating strong generalization and effectiveness.
Challenges remain for ambiguous visuals or overlapping actions with similar semantics, where precise boundaries are difficult. Future work could leverage finer motion cues or stronger temporal constraints to address these limitations.



\bibliographystyle{IEEEtran}
\bibliography{references}

\end{document}